\documentclass{article}
\usepackage{ijcai21}
\usepackage{times}
\usepackage{soul}
\usepackage{url}
\usepackage[hidelinks]{hyperref}
\usepackage[utf8]{inputenc}
\usepackage[small]{caption}
\usepackage{graphicx}
\usepackage{amsmath}
\usepackage{amsthm}
\usepackage{subcaption}
\usepackage{booktabs}
\usepackage{algorithm}
\usepackage{algorithmic}
\usepackage{latexsym}
\usepackage{tikz}
\definecolor{lightblue}{HTML}{38D5FF}
\definecolor{lightgreen}{HTML}{76DA76} 
\definecolor{darkred}{HTML}{AC3333} 
\definecolor{lightgray}{HTML}{B8B8B8} 
\definecolor{darkgray}{HTML}{686868} 
\newcommand{\mysquare}[1]{\tikz{\filldraw[draw=#1,fill=#1] (0,0) rectangle (0.6em,0.6em);}}
\newcommand{\mycircle}[1]{\tikz{\filldraw[draw=#1,fill=#1] (0,0) circle [radius=0.3em];}}

\title{Hack The Box: Fooling Deep Learning Abstraction-Based Monitors}

\author{
Sara Hajj Ibrahim$^1$ \and
Mohamed Nassar$^{1,2}$\\
\affiliations
$^1$ American University of Beirut (AUB)\\
$^2$ University of New Haven \\
\emails
 sih11@mail.aub.edu,
mnassar@newhaven.edu
}

\begin{document}
\maketitle
\begin{abstract}
Deep learning is a type of machine learning that adapts a deep hierarchy of concepts. Deep learning classifiers link the most basic version of concepts at the input layer to the most abstract version of concepts at the output layer, also known as a class or label. However, once trained over a finite set of classes, some deep learning models do not have the power to say that a given input does not belong to any of the classes and simply cannot be linked. Correctly invalidating the prediction of unrelated classes is a challenging problem that has been tackled in many ways in the literature. Novelty detection gives deep learning the ability to output "do not know" for novel/unseen classes. Still, no attention has been given to the security aspects of novelty detection. In this paper, we consider the case study of abstraction-based novelty detection and show that it is not robust against adversarial samples. Moreover, we show the feasibility of crafting adversarial samples that fool the deep learning classifier and bypass the novelty detection monitoring at the same time. In other words, these monitoring boxes are hackable. We demonstrate that novelty detection itself ends up as an attack surface.
\end{abstract}

\section{Introduction}
Machine learning algorithms are excellent at analyzing data and finding interesting patterns. However, they give up to the so-called dimensionality curse. It was shown that deep learning bypasses the traditional machine learning algorithms in most learning tasks in the literature \cite{goodfellow2016deep,nassar-dlh-2020}. While deep learning yields remarkable results in the field of raw data representation and classification, it suddenly becomes sub-optimal when explaining decisions or recognizing a novel class of input.  
In their default settings, supervised deep neural networks never say "I don't know", they can be just less or more confident about an outcome or decision. The necessity of monitoring deep learning for novelty and anomaly detection is directly visible. 

Novelty detection can play a significant role and be leveraged for monitoring and discovering new classes that were unseen during training time. 
However, most work on novelty detection give no attention to its security aspects. In this paper, we show that from a security perspective, novelty detection can be easily attacked and may augment the attack surface of deep learning based systems. 

We distinguish between three interrelated and very close concepts, namely anomaly detection, outlier detection and novelty detection. These terms are sometimes interchangeably used in literature, but we suggest that they mean different things and it is time to give each an appropriate definition. In our terminology, anomaly detection stems from unsupervised one-class modeling or supervised binary classification into normal and abnormal. Outlier detection stems from unsupervised learning and consists on finding points that likely do not belong to any clusters found in unlabeled data. 

Novelty detection is the process of distinguishing between data inputs belonging to one of the classes encountered during the training time and data inputs belonging to classes that are previously unseen. It is different than outlier detection since data have labels. It is also different from anomaly detection since it is parameterized by both the data and the classifier whereas anomaly detection is usually solely parameterized by the data. 
Novelty detection in deep learning is a new and active research area. Abstraction-based novelty detection is one of the main proposed approaches. This approach summarizes training input and intermediate data representations into statistical constructs that makes it easy to detect novelty at the testing stage. 

A white-box abstraction-based novelty detection method is proposed in \cite{outsidethebox19}.
A monitor takes a one-by-one decision on each testing sample and identifies it as either valid (i.e. the classifier prediction is correct on assigning the sample to one of the training classes), or invalid (i.e. the classifier prediction is rejected). The monitor decision is based on verifying whether a special representation of the sample falls within one of the boxes constructed during training or outside these boxes. This special representation is based on values taken from internal neural nodes at hidden layers. Each class has its own box or set of boxes. We take these monitors as a case-study in this paper and show that: (1) these monitors are not efficient when adversarial testing samples are presented, and (2) these monitors can themselves be attacked by appending the adversarial generation process with new constraints. In other words, these boxes are hackable. 

The remaining of this paper is organized as follows: Section 2 summarizes our terminology and literature review. Our attack methodology is presented in section 3. Section 4 evaluates our experiments and findings. Finally Section 5 concludes the paper and sketches future work. 

\section{Background \& Literature Review}
\begin{figure}[tbp] 
\centering
\includegraphics[width=3.3in]{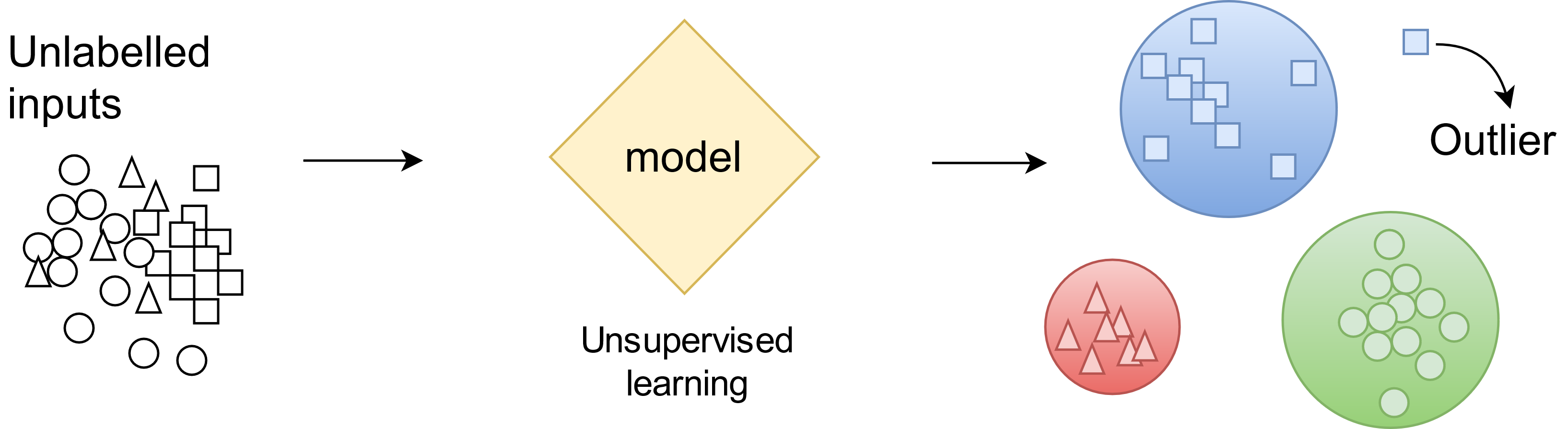}
\caption{An example of an outlier detection system. \label{outlierdetection}} 
\end{figure}

First let's be more precise about what is novelty detection and how it differs from outlier and anomaly detection. 
An outlier is an "Observation which deviates so much from other observations as to arouse suspicion it was generated by a different mechanism" \cite{hawkins1980identification}. 
Outlier detection is the process of coining observations that significantly deviate from the majority of data. Unsupervised algorithms extract statistical information indicating how unlikely a certain observation is to occur, for example, finding a point deviating far from the statistical means of other points as illustrated in Figure \ref{outlierdetection}.

\begin{figure}[tbp] 
\centering
\includegraphics[width=3.3in]{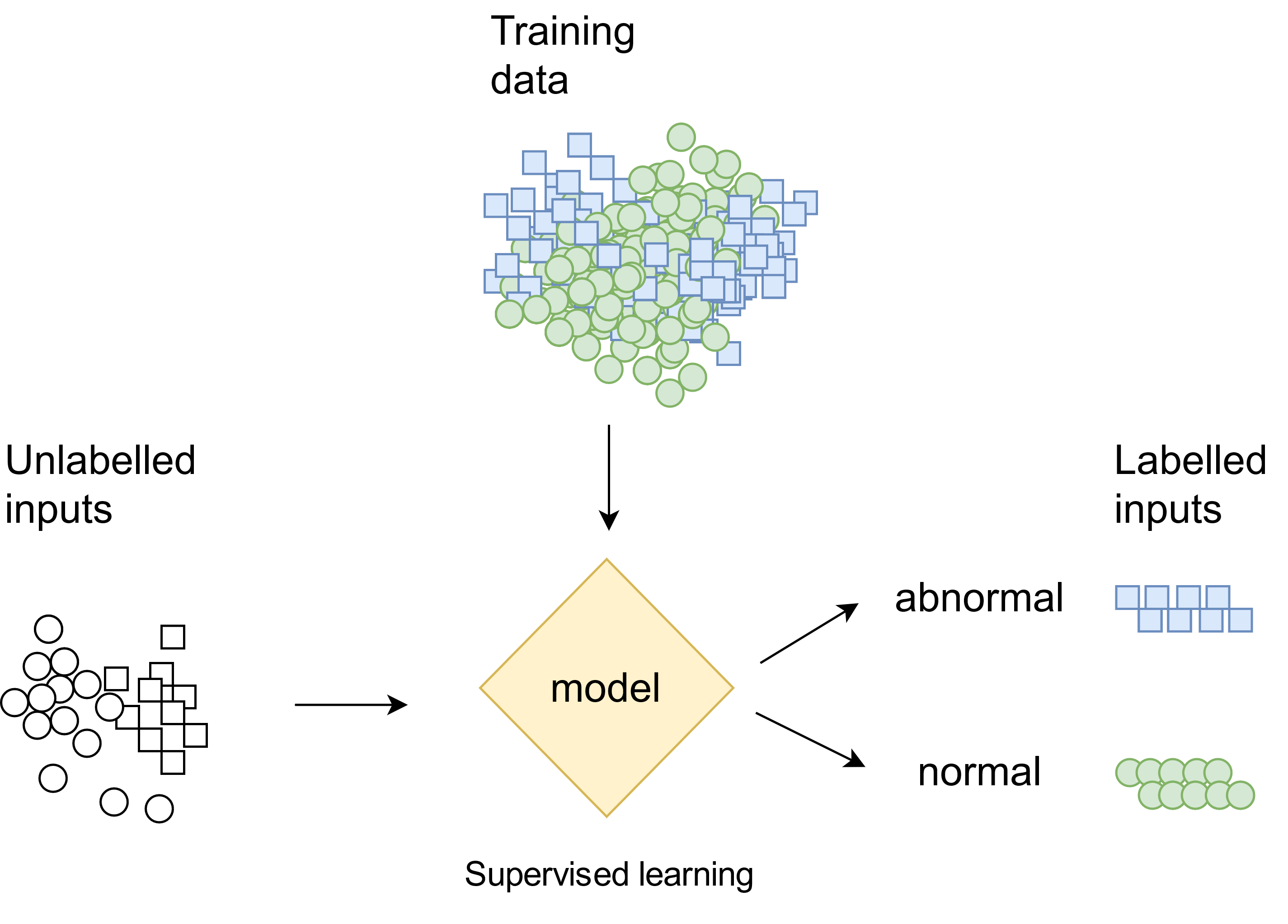}
\caption{An example of an anomaly detection system. \label{anomalydetection}} 
\end{figure}

An anomaly is a special case of outliers which is usually tied to special information or reasons \cite{Aggarwal2015}. Anomalies indicate significant and rare events that may prompt critical actions in a wide range of application domains \cite{ahmed2016survey}. Anomaly detection may require labeled data and employ supervised algorithms as illustrated in Figure \ref{anomalydetection}. For example, we consider the problem of malware/benign classification as a form of anomaly detection. 

Novelty detection is the process of identifying inputs that belong to unknown classes that were not provided during training time. Consider a supervised learner having $c$ classes at training time but $c+u$ classes appear at testing time. The goal of novelty detection is to invalidate the output of the classification when samples from the $u$ classes are presented. Novelty detection is different than the previously described anomaly and outlier detection for two main reasons: (1) training data have labels, and (2) the learner itself is an input to the detector algorithm. 

\begin{figure}[tbp] 
\centering 
  \subfloat[Abstraction phase]{
    \centering 
   \includegraphics[width=3.3in]{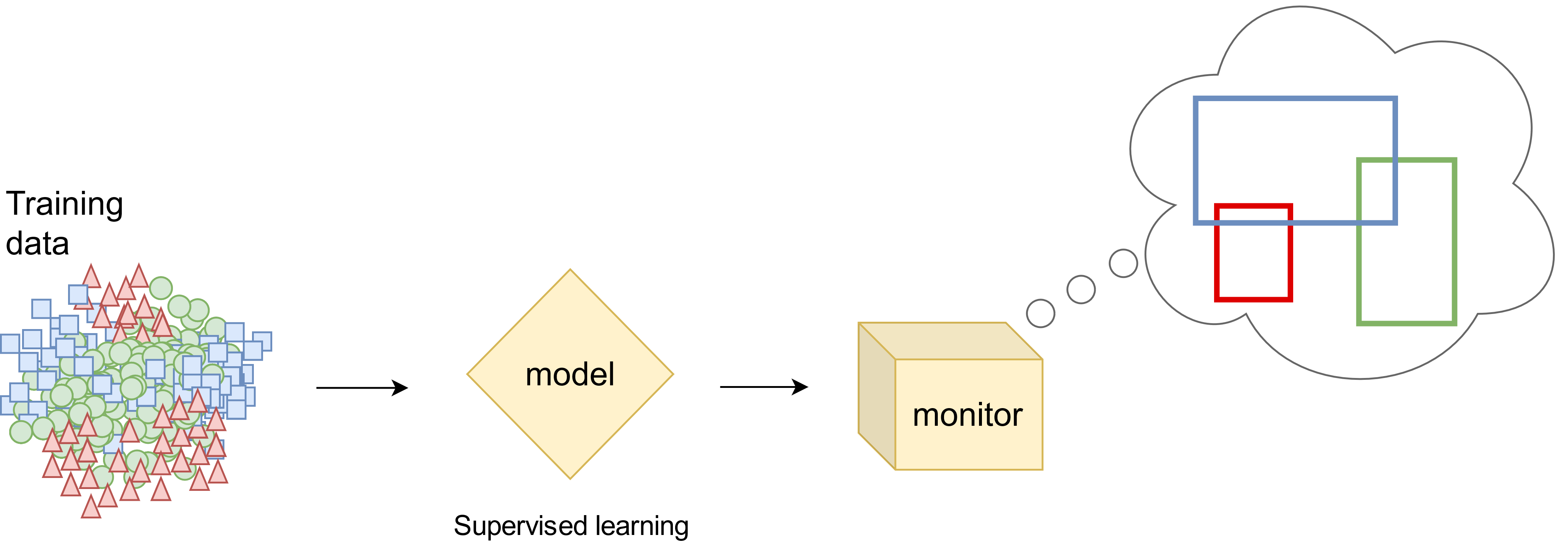}}\\
  \subfloat[Monitoring phase]{
    \centering 
   \includegraphics[width=3.3in]{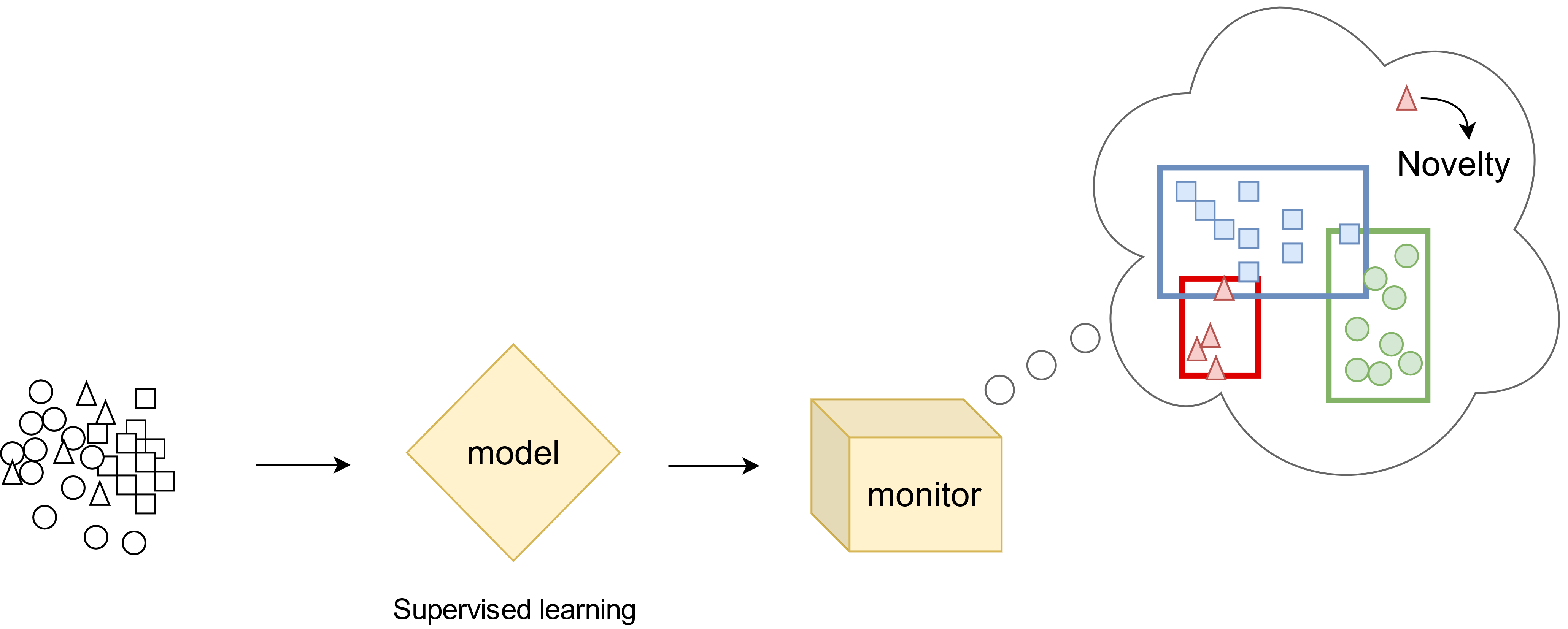}}
   \caption{An example of a novelty detection system based on box abstractions. \label{noveltydetection}}
   
\end{figure}

Novelty detection can be achieved in white-box mode by taking the model obtained after training and building a monitor on top of it. The monitor fingerprints the behaviour of the model when training data are presented. For the special case of deep learning, such a monitor can register the values of hidden nodes given by forward-propagating the training samples. The monitored values are abstracted into statistical constructs. Later, outlier detection flag inputs having fingerprints that largely deviate from these constructs. In other words, this approach transforms the novelty detection problem into an outlier detection problem by projecting the data into a new hyper-dimensional space. This projection is parameterized by the neural network model itself. Different types of abstractions were proposed in \cite{outsidethebox19} before concentrating on the evaluation of box abstractions in particular. Figure \ref{noveltydetection} shows an example of a box abstraction-based novelty detection system. 

Next, we summarize the main approaches for addressing novelty detection in deep learning from the literature. 
\begin{description}

\item  [Distance based methods] These methods compute novelty values or confidence scores based on distance metric functions. In \cite{mandelbaum2017distance} the data are first embedded as derived from the penultimate layer of the neural network. The confidence score is based on the estimation of local density. Local density at a point is estimated based on the Euclidean distance in the embedded space between the point and its $k$ nearest neighbors in the training set. A similar approach based on learning a local model around a test sample is proposed on \cite{bodesheim2015local} for Multi-class novelty detection tasks in image recognition problems. 

\item [Statistical Based Methods] Novelties are caused by differences in data distributions at training and prediction time. Some of these methods require sampling the distribution at run-time or an online adaptation of classifiers. In \cite{pidhorskyi2018generative} the underlying structure of the inlier distribution of the training data is captured. The novelty is detected by the means of a hypothesis test or by computing a novelty probability value. 

\item[Auto-Encoding and Reconstruction Based Methods] One way to proceed is to train a deep encoder-decoder network that outputs a reconstruction error for each sample. The error is used to either compute a novelty score or to train a one-class classifier. \cite{pidhorskyi2018generative} also uses an auto-encoder network but to derive a linearized manifold representation of the training data. The manifold representation helps compute a novelty probability that represents how likely it is that a sample was generated by the inlier distribution. This is why we consider it as a statistical based method in the same time. 

\cite{domingues2018deep} introduces an unsupervised model for novelty detection based on Deep Gaussian Process Auto-Encoders (DGP-AE). The proposed auto-encoder is trained by approximating the DGP layers using random feature expansions, and by performing stochastic variational inference on the resulting approximate model. Their work can be categorized under anomaly detection in our terminology.

\item [Bayesian methods] These methods use Bayesian formalism to detect anomalies and new classes in addition to classification \cite{roberts2019bayesian}. The basic idea is to add a "dummy" class at the root node. The class is considered under-represented in the training set. The classifier gives a strong a posterior of being "dummy" for unseen instances. 

\item [Abstraction based methods] 
These methods consider a finite set of vectors $X$, and construct a set $Y$ that generalizes $X$ to infinitely many elements and has a simple representation that is easy to manipulate and answer queries for. Examples of these methods are ball-abstraction such as one-class support vector machines, one-class neural networks \cite{chalapathy2018anomaly} and box-abstraction \cite{outsidethebox19}.
\end{description}

However, little or none work has been conducted on the security of the aforementioned approaches either in the presence of adversarial samples for fooling the classifier or against especially crafted samples for fooling the novelty detector and the classifier at the same time. For our study, we consider the use-case of "outside the box" \cite{outsidethebox19} and analyse its security in several ways.

\section{Methodology}
In \cite{outsidethebox19}, constructing an abstraction at layer $l$ of the monitored network for class $y$  works as follows: 
\begin{enumerate}
    \item Collect outputs at layer $l$ for inputs of class $y$.
    \item Divide collected vectors into clusters.
    \item Construct an abstraction for vectors in each cluster, e.g. an enclosing box.
\end{enumerate}
Monitoring at layer $l$ works as follows: 

\begin{enumerate} 
\item Predict class of input $\mathbf{x}$.
\item Collect output at layer $l$ into a vector $\mathbf{v}$.
\item Check if any of the abstraction of the predicted class contains $\mathbf{v}$. 
\item The prediction is rejected if the check returns empty. 
\end{enumerate}

We distinguish two types of attacks against this schema: 
\begin{description}
\item[Attack 1 - from valid to invalid] Consider an input $\mathbf{x}$ which would normally be identified as valid by the monitor and belongs to one of the training classes. This attack modifies $\mathbf{x}$ in a slight and unperceivable way to make it get rejected by the monitor. As an example of application of this attack, we would imagine a denial of service for a legitimate user in a face recognition system. 

\item[Attack 2 - from invalid to valid] Consider an input $\mathbf{x}$ which would normally be rejected by the monitor as not belonging to any of the classes seen during training. This attack modifies $\mathbf{x}$ in a slight and unperceivable way to make it get accepted by the monitor. The attack can be targeting a preset prediction, or just going with any prediction output by the neural network. As an example of  application, we would imagine letting go an intruder in a face recognition system. The intruder is identified as any of the legitimate white-list users.
\end{description}

In terms of implementation, we propose to formulate each attack as an optimisation problem that can be solved by an iterative optimisation algorithm. We adapt an idea that has been widely studied in the literature (e.g. \cite{zugner2018adversarial}): introducing a perturbation within a small epsilon-ball or budget to maximize the confusion of the target function.
We also experiment with off-the-shelf adversarial attacks against neural networks and assess their efficiency, as well as augmenting them by an optimisation component to target a specific attack and make them more efficient. We detail these approaches next. 

\subsection{Optimisation based attacks}  
\subsubsection*{Attack 1: from valid to invalid}  

\begin{figure}[t] 
\centering 
  \subfloat[Attack 1: from valid to invalid. The blue squares and green circles are pushed outside the boxes]{
    \centering 
   \includegraphics[width=2.3in]{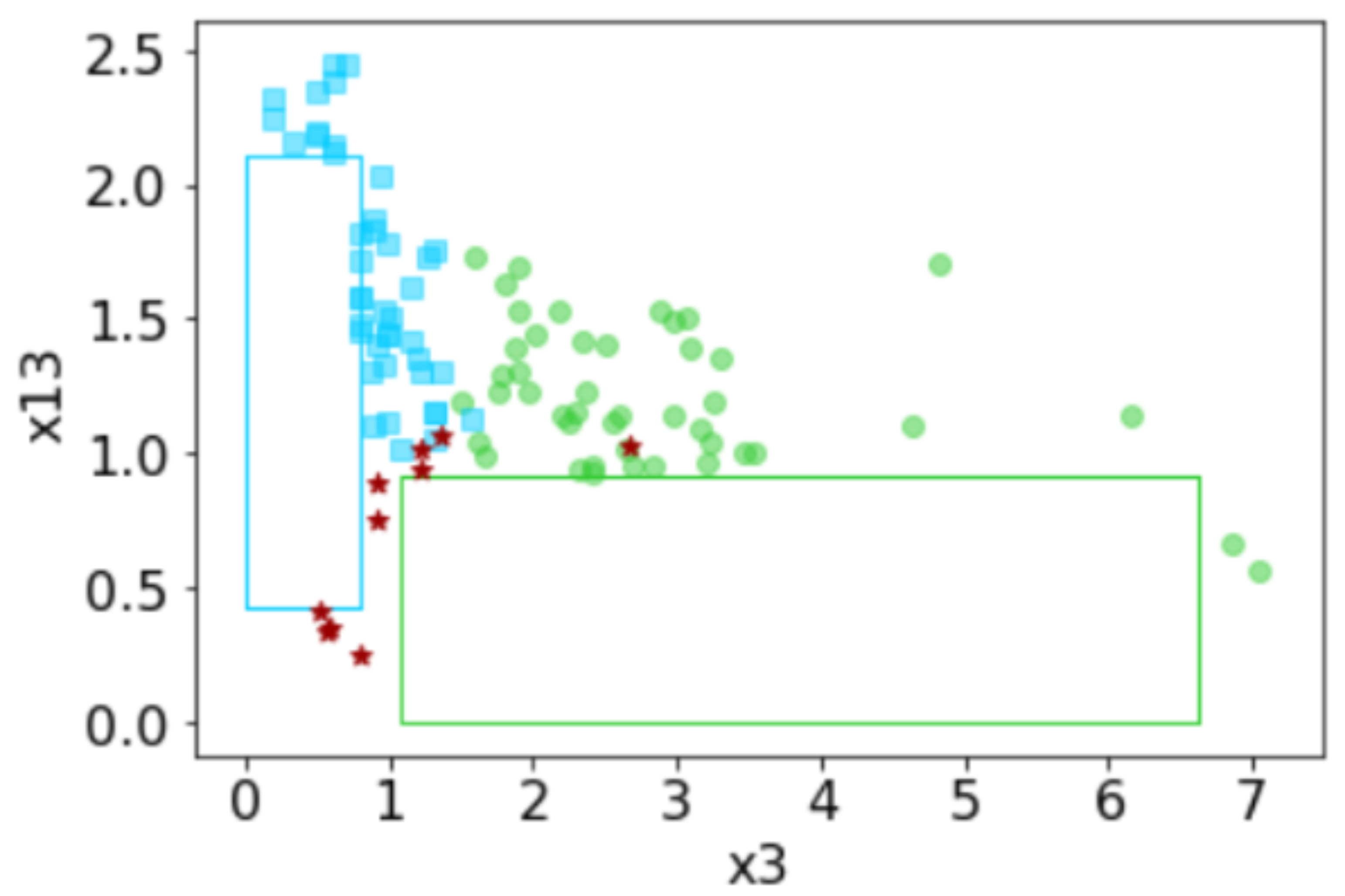}}\\
  \subfloat[Attack 2: from invalid to valid. The red stars are pulled inside the boxes.]{
    \centering 
   \includegraphics[width=2.3in]{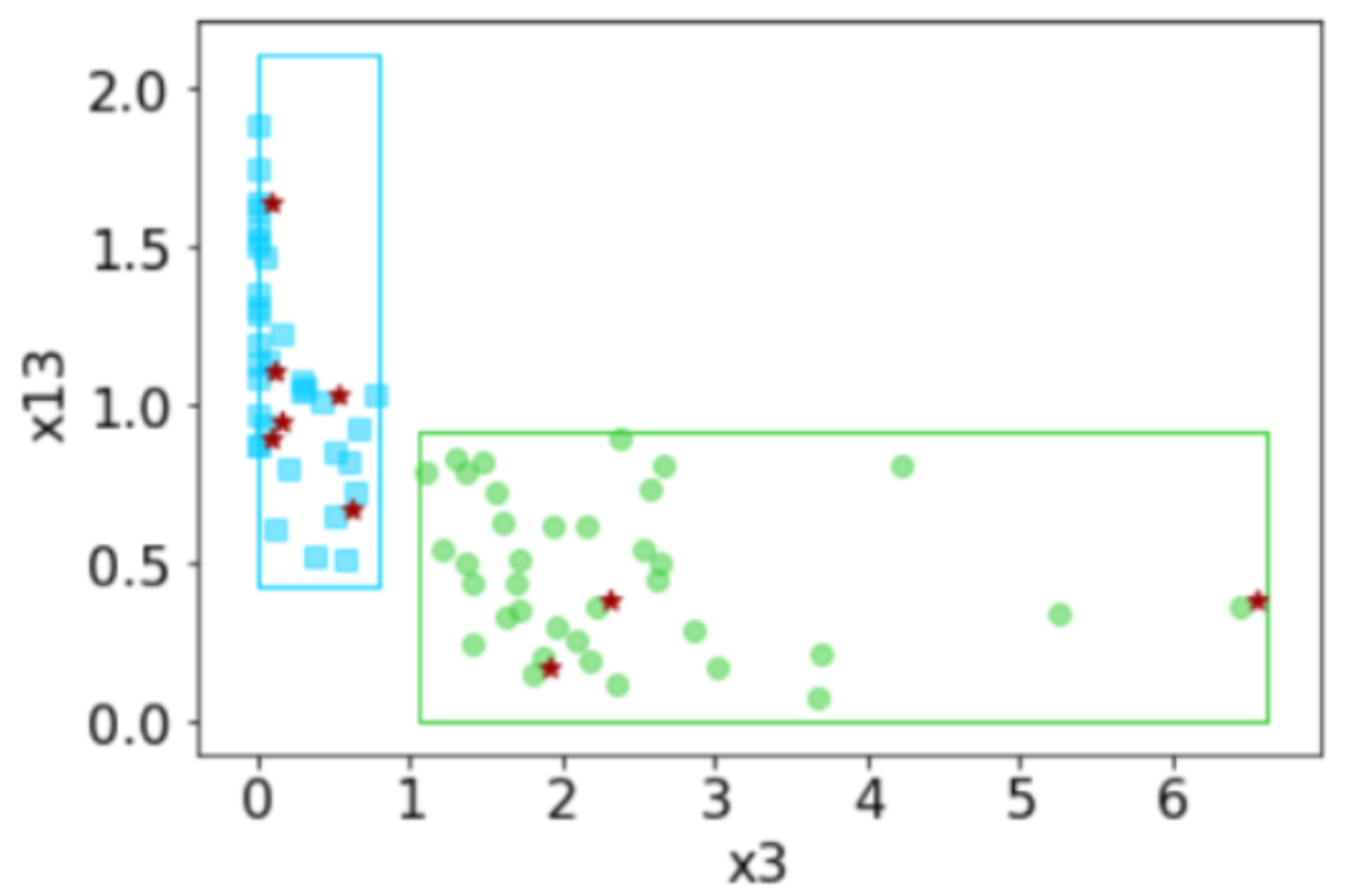}}
   \caption{Examples of attack types. Projections using two arbitrary dimensions are shown.}
   \label{attack_boxes}
\end{figure}

Consider a neural network that is trained over only two classes, where each class is represented by the monitor under one box. As shown in Figure \ref{attack_boxes}($a$), we push the images of valid points, as represented by the monitor, from both classes \mysquare{lightblue} and \mycircle{lightgreen} to fall outside their boxes and therefore be marked as novel exactly as for the \textcolor{darkred}{$\boldsymbol{\star}$} points. 

More generally, consider a point $\mathbf{x^0}$ such as $\text{monitor}(\mathbf{x^0}, c) = 1$ (accept as class $c$), our goal is to find a point $\mathbf{x}$ as close as possible to  $\mathbf{x^0}$ such that $\text{monitor}(\mathbf{x}, c) = 0$ (reject). For measuring the distance between the two points we either use the  $L_1$ norm or the $L_2$ norm. In addition, we require that $\mathbf{x}$ preserves the same prediction as of $\mathbf{x^0}$ via the monitored network:  $\text{predict}(\mathbf{x})=\text{predict}(\mathbf{x^0}) = c$. 

In case of the $L_1$ norm, we replace the non-differentiable objective function by a differentiable one through introducing a vector $\mathbf{z}$ having the same dimension as $\mathbf{x}$. We get a linear programming problem as follows: 

\begin{equation}
\begin{aligned}
&\text{Minimize} & & \sum_{i=1}^{n}z_i \\
&\text{Subject to} & & z_i + (x_i - x_i^0) \geq 0  \\
&& & z_i - (x_i - x_i^0) \geq 0 \\
&& & \text{monitor}(\mathbf{x}, \text{predict}(\mathbf{x})) = 0 \\
&& & \text{predict}(\mathbf{x}) = \text{predict}(\mathbf{x^0})  \\
\end{aligned}
\end{equation}
 
We next use the $L_2$ norm (${\left\|\mathbf{x} - \mathbf{x^0} \right\|}_2$) to formulate a second optimisation problem. The constraints for $L_2$ are same as for $L_1$ except that $L_2$ norm is already a differential objective function and can be directly tuned by a linear solver. The $L_2$ norm attack focuses on minimizing the square root of the sum of squared differences of ($\mathbf{x^0}$) and ($\mathbf{x}$) elements. We then get a linear programming problem as follows:

\begin{equation}
\begin{aligned}
&\text{Minimize} & & \sqrt{\sum_{i=1}^{n}{|x_i - x_i^0|}^2} \\
&\text{Subject to} & & \text{monitor}(\mathbf{x}, \text{predict}(\mathbf{x})) = 0 \\
&& & \text{predict}(\mathbf{x}) = \text{predict}(\mathbf{x^0})  \\
\end{aligned}
\end{equation}

\subsubsection*{Attack 2: from invalid to valid}
In this attack, we push the images of invalid points \textcolor{darkred}{$\boldsymbol{\star}$}, as represented by the monitor, towards the boxes representing legitimate classes. 
The points will be marked by the monitor as valid if, at the same time, the neural network prediction matches the box owner, either as \mysquare{lightblue} or \mycircle{lightgreen}. The 2d projection of the monitor representation for a binary classifier is shown in Figure \ref{attack_boxes}($b$).  

Given a point $\mathbf{x^0}$ where $\text{monitor}(\mathbf{x^0}, c) = 0$ (reject), our goal is to find a point $\mathbf{x}$ as close as possible to $\mathbf{x^0}$ such that $\text{monitor}(\mathbf{x}, \text{predict}(\mathbf{x})) = 1$ (accept as same class of $\mathbf{x^0}$). For measuring the distance between the two points we choose the $L_1$ norm or the $L_2$ norm. 

For the $L_1$ norm, we formulate the following problem: 
\begin{equation}
\begin{aligned}
&\text{Minimize} & & \sum_{i=1}^{n}z_i \\
&\text{Subject to} & & z_i + (x_i - x_i^0) \geq 0  \\
&& & z_i - (x_i - x_i^0) \geq 0 \\
&& & \text{monitor}(\mathbf{x}, \text{predict}(\mathbf{x})) = 1 \\
&& & \text{predict}(\mathbf{x}) = \text{predict}(\mathbf{x^0})  \\
\end{aligned}
\end{equation}

For the $L_2$ norm, we formulate the following problem: 
\begin{equation}
\begin{aligned}
&\text{Minimize} & & \sqrt{\sum_{i=1}^{n}{|x_i - x_i^0|}^2} \\
&\text{Subject to} & & \text{monitor}(\mathbf{x}, \text{predict}(\mathbf{x})) = 1 \\
&& & \text{predict}(\mathbf{x}) = \text{predict}(\mathbf{x^0})  \\
\end{aligned}
\end{equation}

The four formulated problems can be efficiently solved by constrained optimisation numerical methods that are either local such as SLSQP \cite{kraft1988software} and COBYLA \cite{powell1994direct}
or global such as Differential Evolution (DE) \cite{price2013differential} and SHGO \cite{endres2018simplicial}. We used implementations from the SciPy library \cite{2020SciPy-NMeth} to show case these attacks as will be detailed later in section 4. 
\subsection{Adversarial attacks against neural networks}
Another idea is to use known adversarial attacks against neural networks as a starting point for attacking the monitor. 
Adversarial neural network attacks aim at changing the prediction of  an input $\mathbf{x^0}$ when replaced by a close point $\mathbf{x}$ as of $\text{predict}(\mathbf{x}) \neq \text{predict}(\mathbf{x^0})$. This can be done in white-box mode by taking a step in the opposite direction of the gradient of the objective function at the point $\mathbf{x^0}$. It is interesting to check whether adversarial samples would be detected as novel by the novelty monitor. Whether should or must they be detected as novel is even a more puzzling question. In case these samples are rejected because of the novelty detector, hence making the attack fail, we may consolidate them by optimisation based methods to make them pass as valid points. In other words, we can take an adversarial sample as a starting point for our search for an attack against both the neural network and the monitor. Our goal here is to find a point $\mathbf{x}$ very close to $\mathbf{x^0}$ such that $\text{predict}(\mathbf{x}) \neq \text{predict}(\mathbf{x^0})$
 and $\text{monitor}(\mathbf{x}, \text{predict}(\mathbf{x})) = 1$ (accept). 

We use the implementation of adversarial attacks from the Foolbox library \cite{rauber2017foolboxnative}. We assess the performance of an abstraction based monitor to flag adversarial samples as novel. We study the effect of tuning "outside the box" parameters to enhance the robustness of the monitor. In a second time, we use optimisation techniques to force adversarial samples to bypass the monitor. 

\begin{figure}[t]
\centering
\includegraphics[width=3.3in]{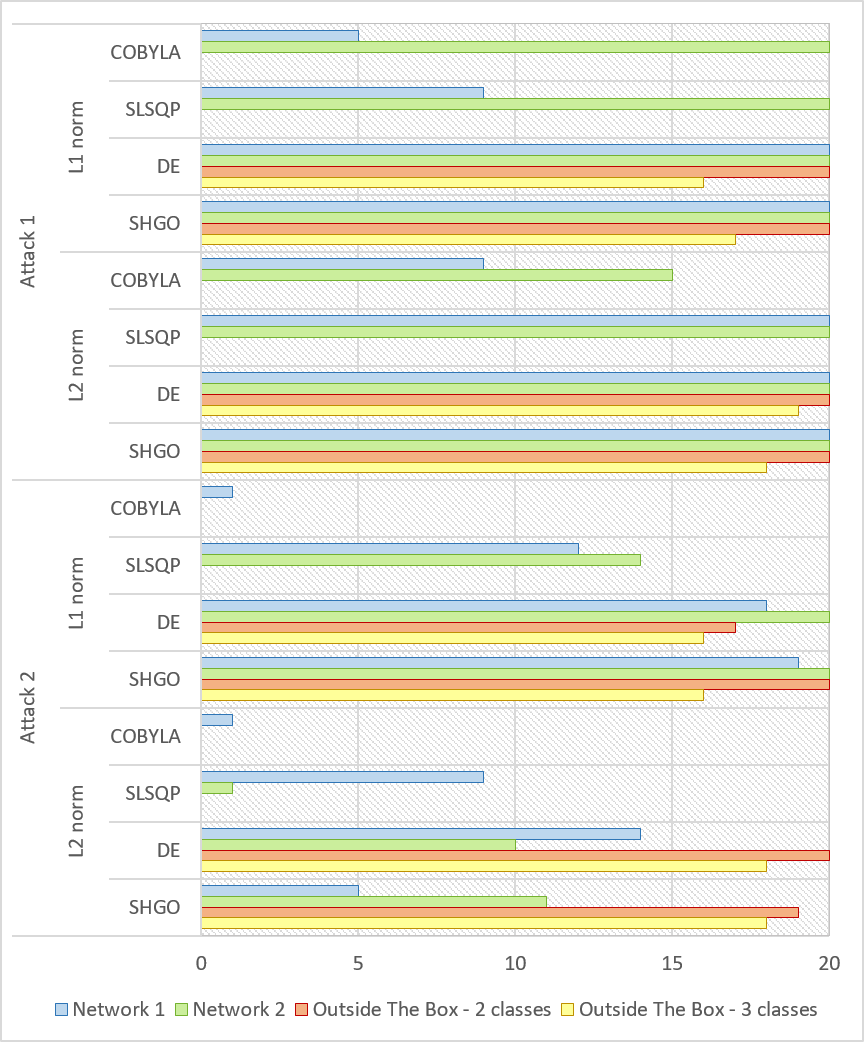}
\caption{Success rate of different optimisation based methods for the four proposed attacks with different network architectures.
\label{optimisation_result}} 
\end{figure} 

\begin{figure}[t]
\centering
 \begin{subfigure}[b]{0.15\textwidth}
\includegraphics[width=\textwidth]{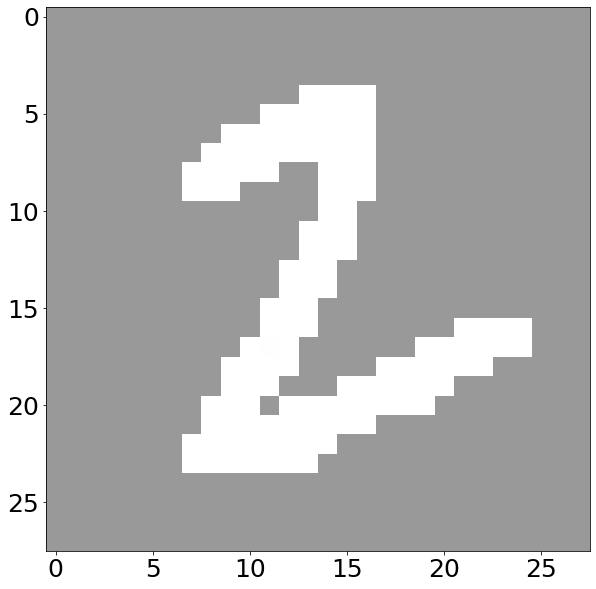}
\caption{Original}
\end{subfigure} \quad \quad
 \begin{subfigure}[b]{0.15\textwidth}
\includegraphics[width=\textwidth]{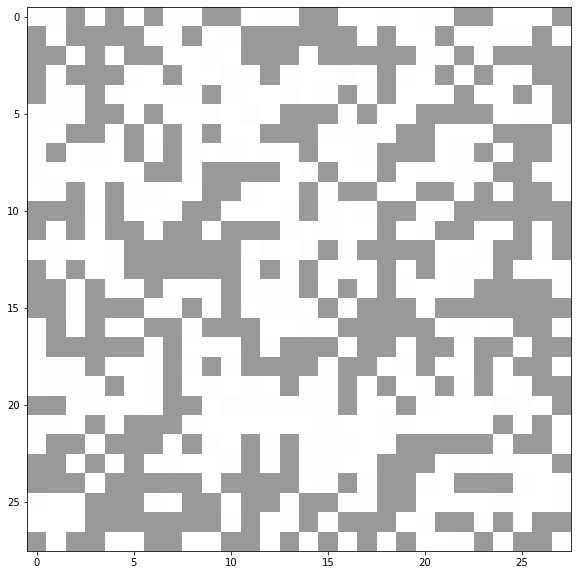}
\caption{Novel}
    \end{subfigure}%
    \caption{Example of an unsuccessful attack / noisy sample obtained by the DE method.}
    \label{noisy_example}
\end{figure}

\section{Experiments} 
\subsection{Optimisation solvers experiment}
In this experiment, we evaluate the performance of different optimisation solvers in solving the four proposed optimisation problems and successfully generating adversarial samples. We evaluate four solvers from SciPy: COBYLA, SLSQP, SHGO and DE. We experiment with several network architectures, the following in particular: 
\begin{itemize}
    \item Two very simple neural networks: a XOR-like circuit and another inspired from \url{https://playground.tensorflow.org/}. 
    \item Two MNIST classifiers trained over a subset of the ten available classes. The first one is trained over two classes, and the eight others are considered novel. The second one is trained over three classes, and the seven others are considered novel. 
\end{itemize}
The novelty monitor follows "outside the box" paradigm. For each network, we tested over 20 random $\mathbf{x^0}$ samples and counted the times where the optimiser found a solution, i.e. generated an effective attack point $\mathbf{x}$.  Results are shown in Figure \ref{optimisation_result}. 

\begin{figure}[tbp]
    \centering
    \begin{subfigure}[b]{.45\textwidth}
    \centering
    \includegraphics[width=1.52cm,height=1.52cm]{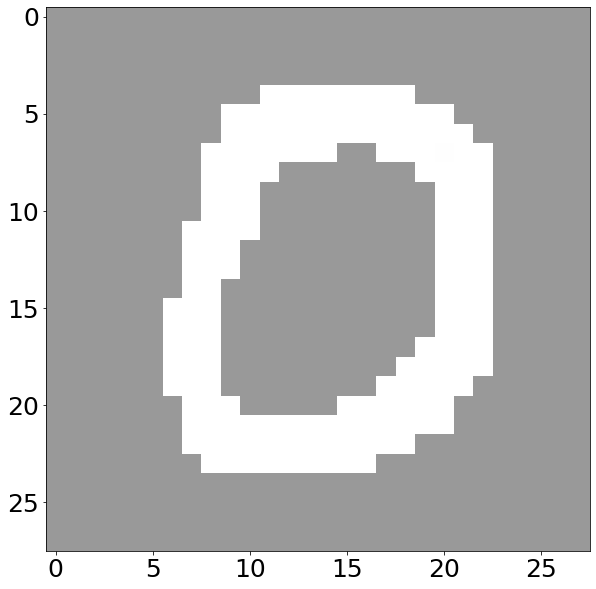}
    \includegraphics[width=1.52cm,height=1.52cm]{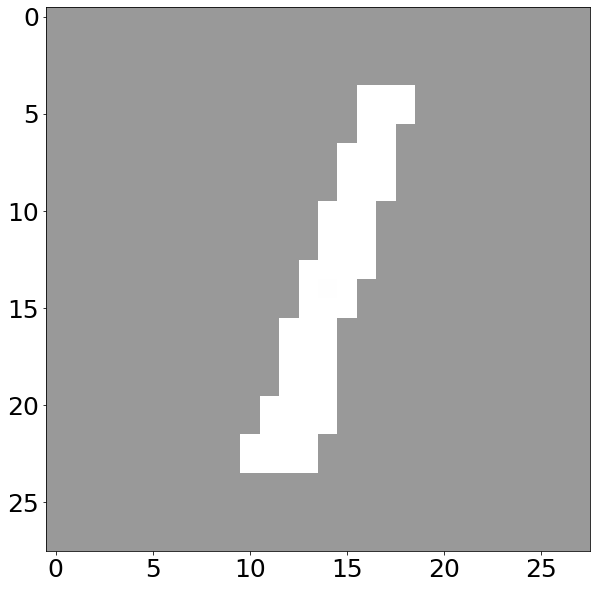}
    \includegraphics[width=1.52cm,height=1.52cm]{images/2_valid.png} 
    \includegraphics[width=1.52cm,height=1.52cm]{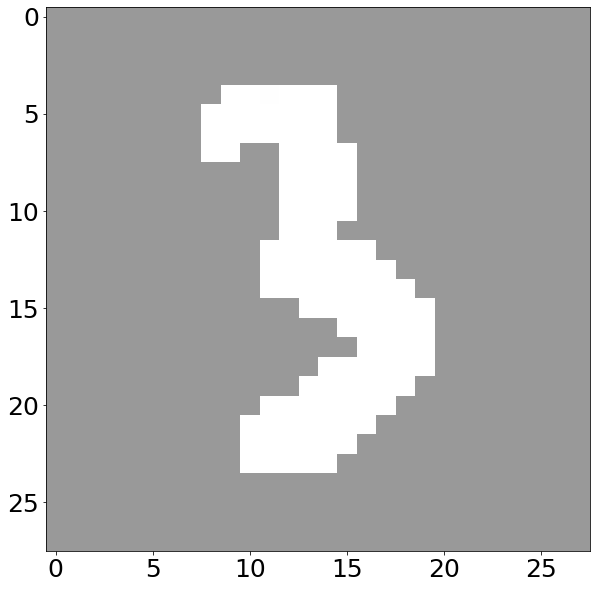} 
    \includegraphics[width=1.52cm,height=1.52cm]{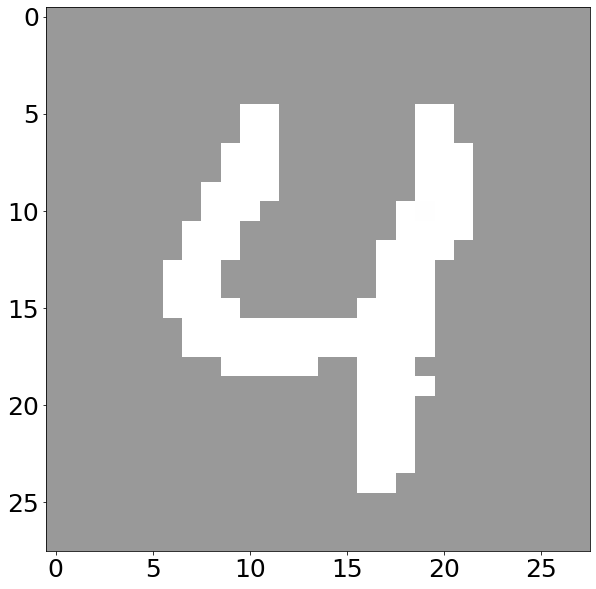} \\
    \includegraphics[width=1.52cm,height=1.52cm]{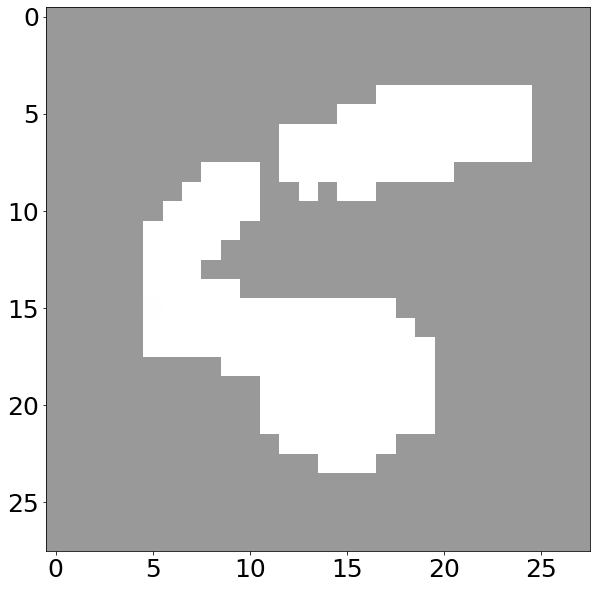}
    \includegraphics[width=1.52cm,height=1.52cm]{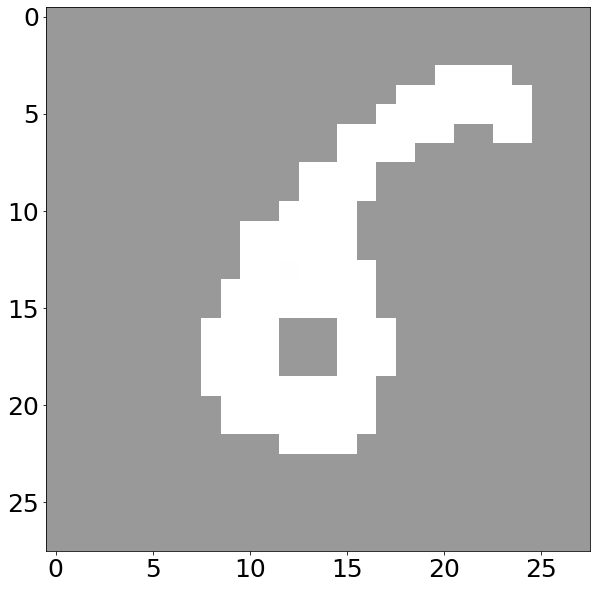}
    \includegraphics[width=1.52cm,height=1.52cm]{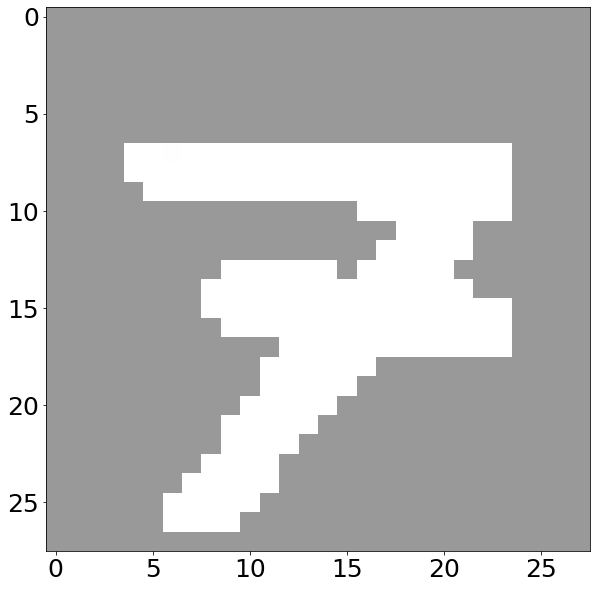} 
    \includegraphics[width=1.52cm,height=1.52cm]{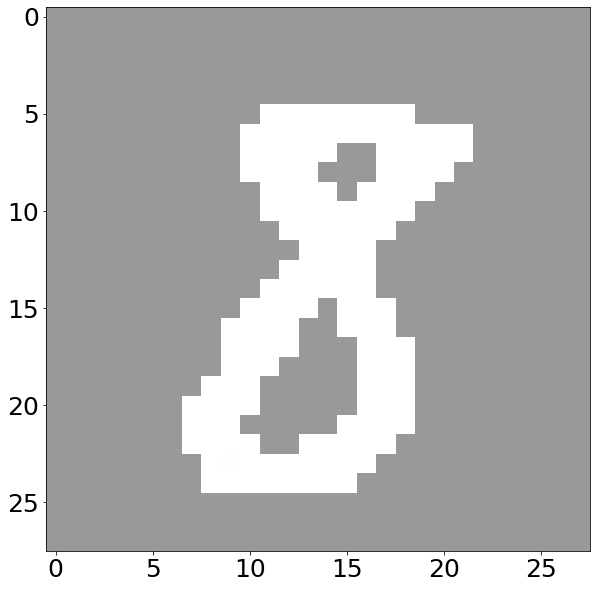} 
    \includegraphics[width=1.52cm,height=1.52cm]{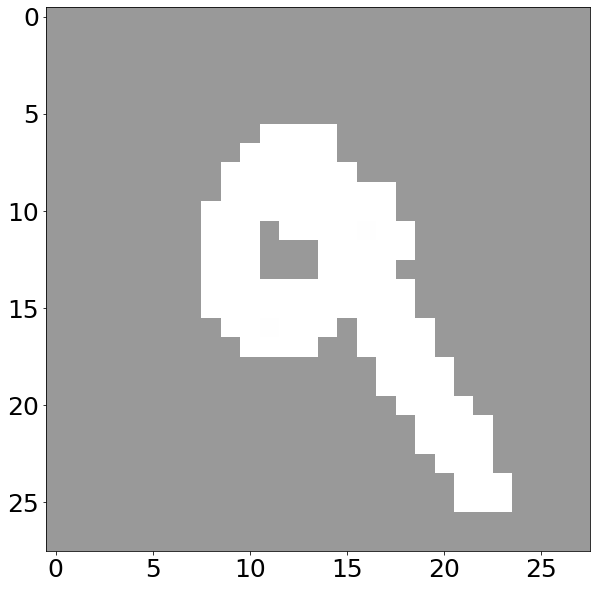} 
    \caption{Before SHGO Optimisation Attack. Samples are decided as not novel.}
    \end{subfigure}%
    
    \begin{subfigure}[b]{.45\textwidth}
    \centering
    \includegraphics[width=1.52cm,height=1.52cm]{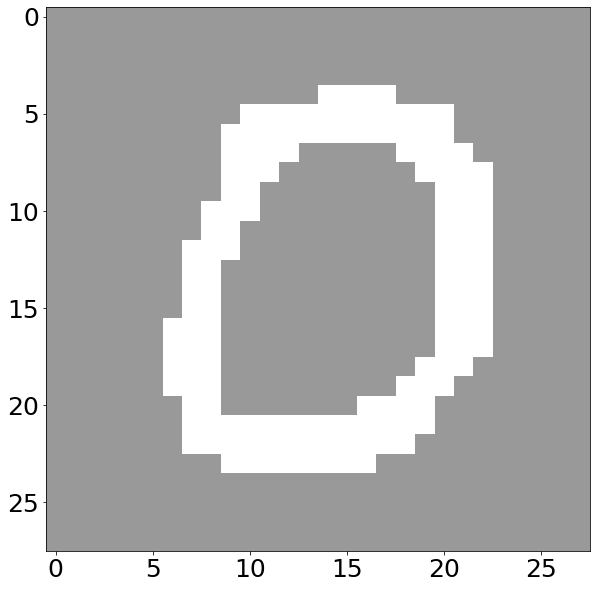}
    \includegraphics[width=1.52cm,height=1.52cm]{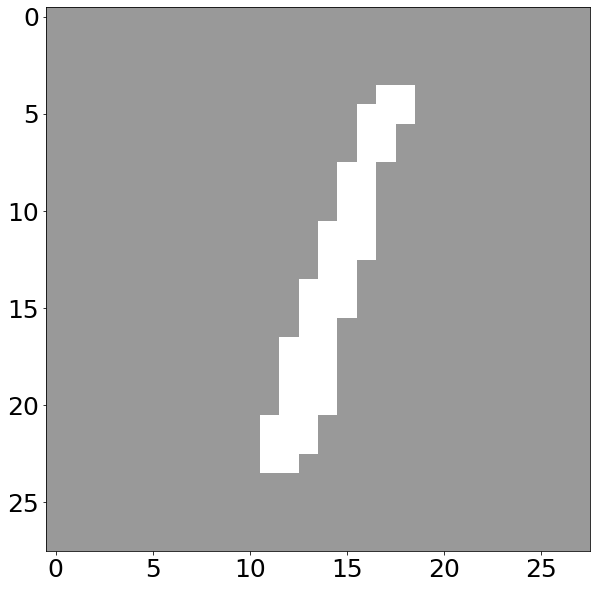}
    \includegraphics[width=1.52cm,height=1.52cm]{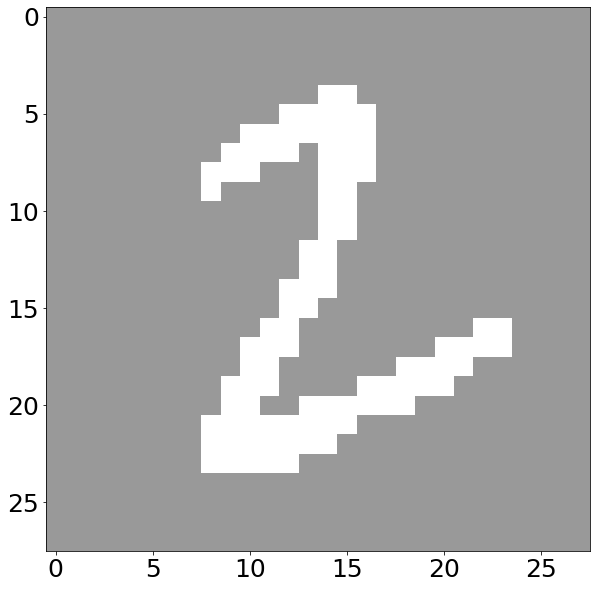} 
    \includegraphics[width=1.52cm,height=1.52cm]{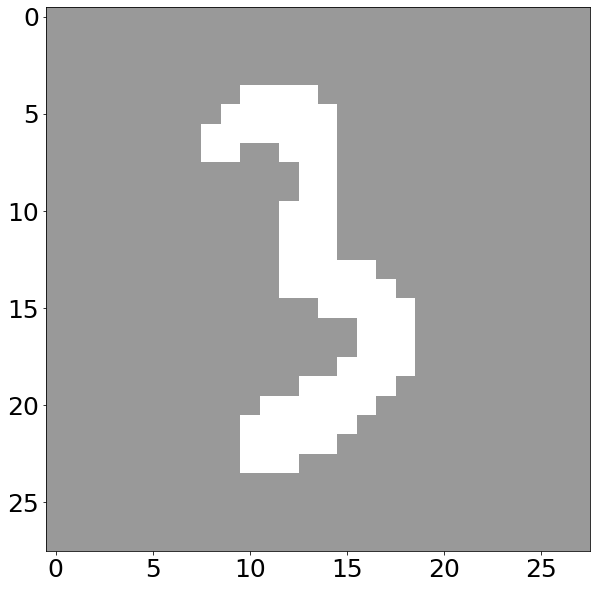} 
    \includegraphics[width=1.52cm,height=1.52cm]{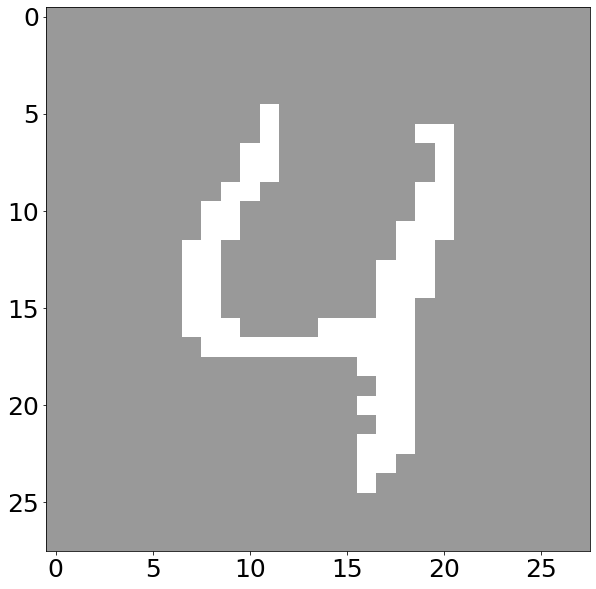} \\
    \includegraphics[width=1.52cm,height=1.52cm]{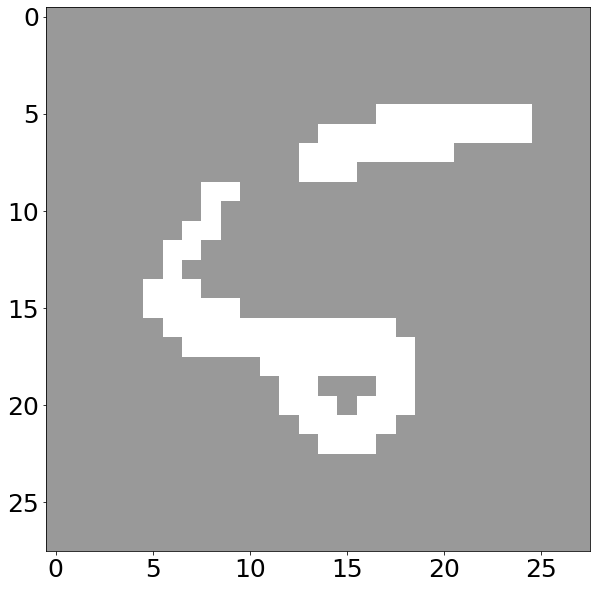}
    \includegraphics[width=1.52cm,height=1.52cm]{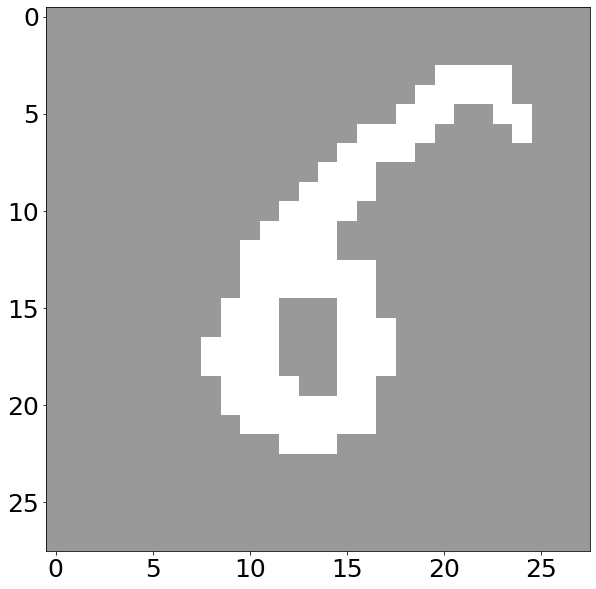}
    \includegraphics[width=1.52cm,height=1.52cm]{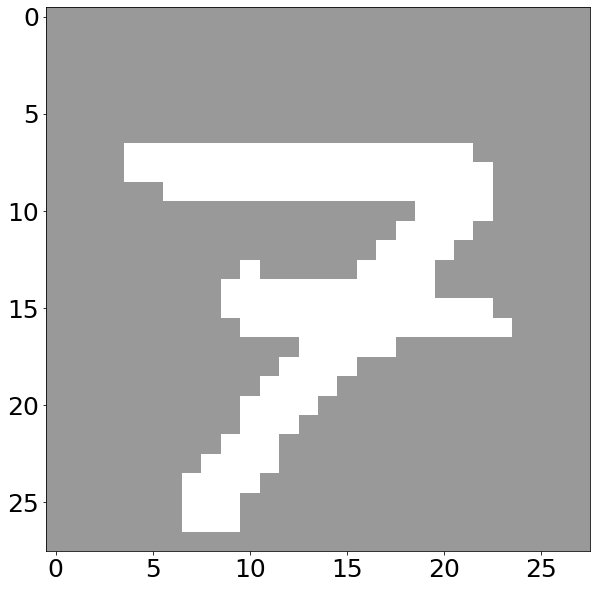} 
    \includegraphics[width=1.52cm,height=1.52cm]{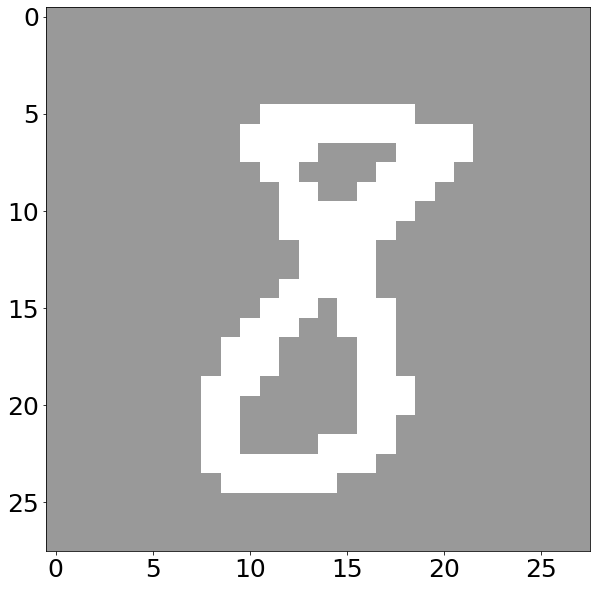} 
    \includegraphics[width=1.52cm,height=1.52cm]{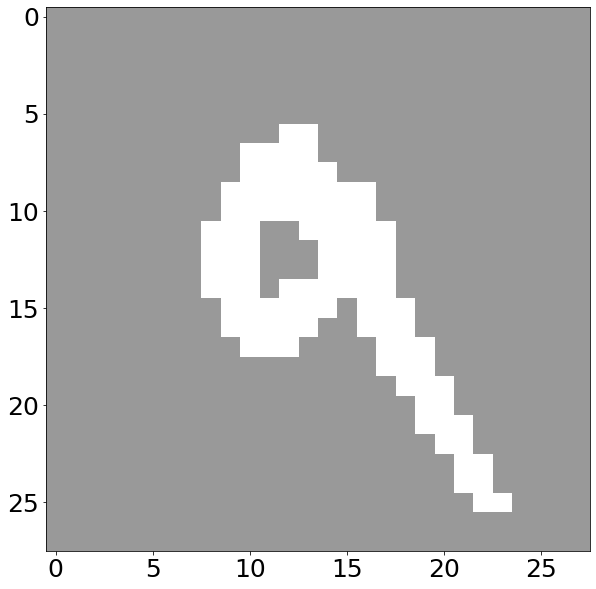} 
    \caption{After SHGO Optimisation Attack. Samples are decided as novel.}
    \end{subfigure}%
    
    \caption{Adversarial image examples obtained by SHGO method.}
    \label{attack_shgo_images}
\end{figure}

\begin{figure}[tbp] 
\centering 
  \subfloat[Attack 1: pushing the valid points outside the boxes.]{
    \centering 
    \includegraphics[width=2.3in]{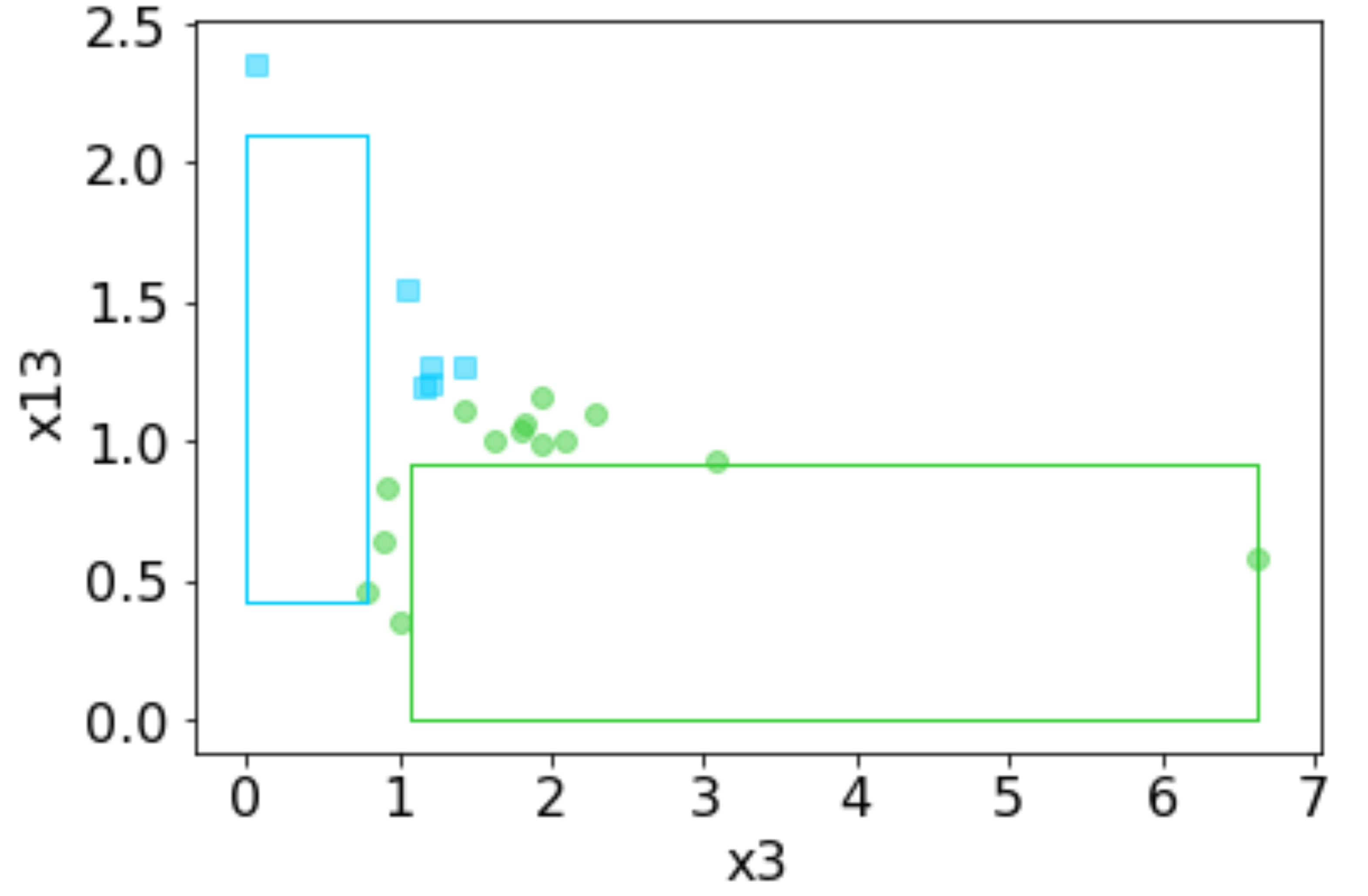}}\\
  \subfloat[Attack 2: pulling the novelties inside the boxes.]{
    \centering 
    \includegraphics[width=2.3in]{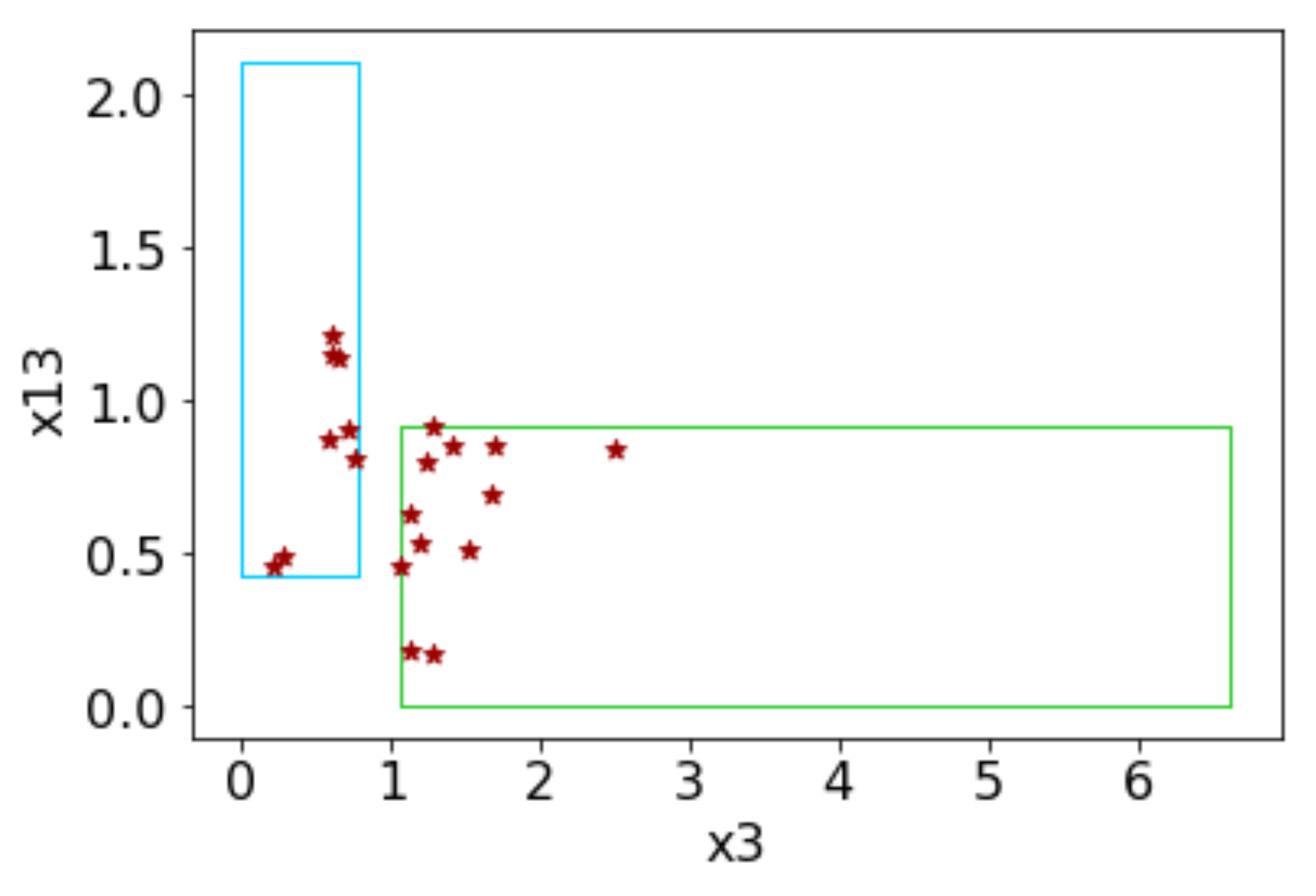}}
    \caption{$L_1$-norm optimisation attacks using SHGO method.}
   \label{Outside_the_Box_attack_2c}
\end{figure}     

Results show that local optimisation methods such as SLSQP and COBYLA were unsuccessful when applied to the MNIST classifiers. We attribute this to the $28*28$ dimensionality of the images. Changing many pixels result in a relatively large distance from the original point, which made the search very narrow around the given starting point. Global optimisation methods (SHGO, DE) were much more successful. 
DE has more successful attempts than SHGO in some cases, but the generated image samples appeared very noisy making the attack easily perceived by a human observer (for an example, see Figure \ref{noisy_example}). SHGO is the best method in terms of success rate and preserving the original digit shape.

Figure \ref{attack_shgo_images} illustrates examples of MNIST images before and after "Attack 1" type (from valid to invalid). The classifier and monitor were trained over the ten classes in this experiment. 

Note that even that our objective function is linear, our constraints such as  $\text{monitor}(\mathbf{x}) = 0/1$ and $\text{predict}(\mathbf{x})=\text{predict}(\mathbf{x^0})$ are strongly non-linear, which hinders the task of the used linear solvers. SHGO shined since it is a derivative-free optimiser that is most appropriate for black box functions and leverages input/output pairs. Another comparison factor is the optimiser runtime. We recorded large run-times ($10-15$ hours) for most optimisers during the experiment using a 16GB RAM computer and a processor of 2.60 GHz frequency. However, SHGO converged in matter of few minutes. Figure \ref{Outside_the_Box_attack_2c} shows a 2d projection of successful generated examples using SHGO and the $L_1$-norm formulation for both types of attacks. 

\subsection{Adversarial attack experiment}
Adversarial attacks on deployments of neural networks are regularly published \cite{ibitoye2019analyzing,aiken2019investigating,papadopoulos2021launching}.
Let us consider the case of adversarial samples that are supposed to fool the classifier decision.
An important question is whether these samples should be flagged as novelty or not. In principle, the goal of novelty detection is not to detect and eradicate adversarial samples.  
It might be reasonable to have the monitor accept the adversarial samples since they are very close to their original counterparts. However, if an adversarial sample is detected as novel, the classifier output will be rejected and the attack will be considered a failure. Therefore, the goal of the attacker becomes to bypass the novelty detection and fool the classifier in the same time.

\begin{figure}[t] 
\centering
\includegraphics[width=3.3in]{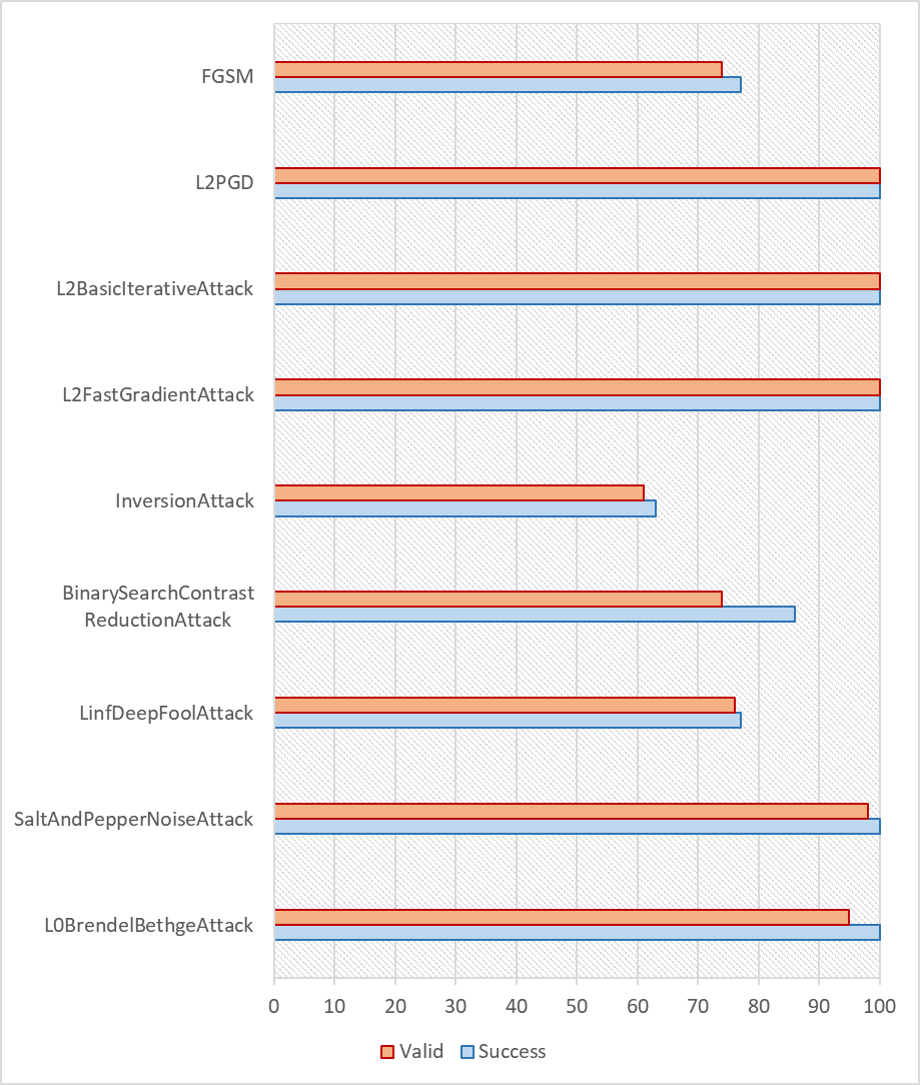}
\caption{Experiment measuring the number of successful and valid attack attempts using neural network attacks.
\label{NN_result}} 
\end{figure}

In this experiment, we consider a neural network classifier trained over all the classes of the MNIST dataset. "Outside the box" novelty detection was not designed with the goal of detecting adversarial samples. However, we check whether such samples would be detected as novel or not. Over 100 MNIST samples, Figure \ref{NN_result} shows the number of successful attacks (i.e., the prediction was successfully flipped) and the number of successful attacks considered as valid (i.e. the monitor validates the prediction considering the sample as not novel). Results show that most of these samples succeeded into fooling the monitor in addition to fooling the classifier. The initial environment properties of the detector were used, hence a single cluster/box for each class and $0$ tolerance. Increasing the number of clusters per class from only one to 2 or 3 clusters per class has a positive impact on invalidating adversarial samples. Conversely, increasing the tolerance factor from zero to $0.1$ or $0.25$ resulted in accepting more adversarial samples. 

\begin{figure}[t] 
\centering 
  \subfloat[Adversarial samples acceptance before optimisation (red stars are rejected).]{
    \centering 
    \includegraphics[width=2.3in]{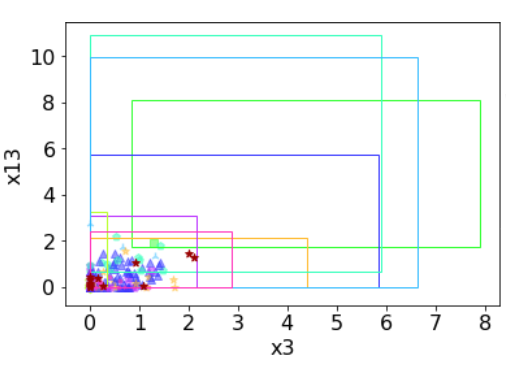}}\\
  \subfloat[Adversarial samples acceptance after optimisation (no red stars).]{
    \centering 
    \includegraphics[width=2.3in]{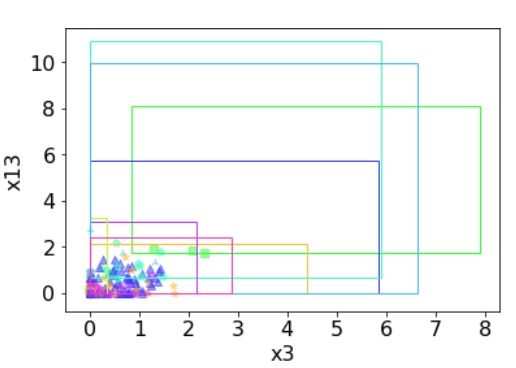}}
    \caption{Fooling the classifier and the monitor together.} 
    \label{NN_opt_Outside_The_Box}
\end{figure}  

Furthermore, the adversarial samples that were successfully detected as invalid can undergo one of our optimisation attacks to pass the monitoring test. For instance, we ran the $L_1$ norm "Attack 2" using SHGO on top of the Fast Gradient Sign Method (FGSM) attack \cite{goodfellow2014explaining} to achieve $100\%$ attack success over 300 MNIST samples. Of course, we had to suppress the $\text{predict}(\mathbf{x}) \neq \text{predict}(\mathbf{x^0})$ constraint. 

Figure \ref{NN_opt_Outside_The_Box} shows a 2d projection of the adversarial samples as per the monitor definition. Before the optimisation attack, there are still adversarial samples marked as \textcolor{darkred}{$\boldsymbol{\star}$} points. After the attack, all \textcolor{darkred}{$\boldsymbol{\star}$} points disappear.

\section{Conclusion and Future Work}

In this paper, we demonstrated that novelty detection monitors are vulnerable to fooling attacks. We were successfully able to mislead the monitor using multiple methods. We formulated optimisation problems that can be solved efficiently to find attack vectors. We also leveraged adversarial neural networks attacks from the literature to fool the classifier and the monitor at the same time.  Adversarial neural network attacks combined with optimisation techniques are shown to be a deadly combo. 

The message of the paper is that security by design should be a requirement for new novelty detection systems in deep learning, especially in critical systems. We envision exploring ways to defend novelty detection against adversarial attacks. In future work, we aim at proposing efficient defense mechanisms for novelty detection monitors against both monitor fooling and classifier-monitor fooling attacks. 

\section*{Acknowledgments}
The authors would like to thank Mohamed Jaber and Yliès Falcone for the introduction and the early discussion of the topic. The authors thank also the anonymous reviewers for their valuable feedback.

\bibliographystyle{named.bst}
\bibliography{main.bib}

\begin{thebibliography}{}

\bibitem[\protect\citeauthoryear{Aggarwal}{2015}]{Aggarwal2015}
Charu~C. Aggarwal.
\newblock {\em Outlier Analysis}, pages 237--263.
\newblock Springer International Publishing, Cham, 2015.

\bibitem[\protect\citeauthoryear{Ahmed \bgroup \em et al.\egroup
  }{2016}]{ahmed2016survey}
Mohiuddin Ahmed, Abdun~Naser Mahmood, and Jiankun Hu.
\newblock A survey of network anomaly detection techniques.
\newblock {\em Journal of Network and Computer Applications}, 60:19--31, 2016.

\bibitem[\protect\citeauthoryear{Aiken and
  Scott-Hayward}{2019}]{aiken2019investigating}
James Aiken and Sandra Scott-Hayward.
\newblock Investigating adversarial attacks against network intrusion detection
  systems in sdns.
\newblock In {\em 2019 IEEE Conference on Network Function Virtualization and
  Software Defined Networks (NFV-SDN)}, pages 1--7. IEEE, 2019.

\bibitem[\protect\citeauthoryear{Bodesheim \bgroup \em et al.\egroup
  }{2015}]{bodesheim2015local}
Paul Bodesheim, Alexander Freytag, Erik Rodner, and Joachim Denzler.
\newblock Local novelty detection in multi-class recognition problems.
\newblock In {\em 2015 IEEE Winter Conference on Applications of Computer
  Vision}, pages 813--820. IEEE, 2015.

\bibitem[\protect\citeauthoryear{Chalapathy \bgroup \em et al.\egroup
  }{2018}]{chalapathy2018anomaly}
Raghavendra Chalapathy, Aditya~Krishna Menon, and Sanjay Chawla.
\newblock Anomaly detection using one-class neural networks.
\newblock {\em arXiv preprint arXiv:1802.06360}, 2018.

\bibitem[\protect\citeauthoryear{Domingues \bgroup \em et al.\egroup
  }{2018}]{domingues2018deep}
R{\'e}mi Domingues, Pietro Michiardi, Jihane Zouaoui, and Maurizio Filippone.
\newblock Deep gaussian process autoencoders for novelty detection.
\newblock {\em Machine Learning}, 107(8-10):1363--1383, 2018.

\bibitem[\protect\citeauthoryear{Endres \bgroup \em et al.\egroup
  }{2018}]{endres2018simplicial}
Stefan~C Endres, Carl Sandrock, and Walter~W Focke.
\newblock A simplicial homology algorithm for lipschitz optimisation.
\newblock {\em Journal of Global Optimization}, 72(2):181--217, 2018.

\bibitem[\protect\citeauthoryear{Goodfellow \bgroup \em et al.\egroup
  }{2014}]{goodfellow2014explaining}
Ian~J Goodfellow, Jonathon Shlens, and Christian Szegedy.
\newblock Explaining and harnessing adversarial examples.
\newblock {\em arXiv preprint arXiv:1412.6572}, 2014.

\bibitem[\protect\citeauthoryear{Goodfellow \bgroup \em et al.\egroup
  }{2016}]{goodfellow2016deep}
Ian Goodfellow, Yoshua Bengio, and Aaron Courville.
\newblock {\em Deep learning}.
\newblock MIT press, 2016.

\bibitem[\protect\citeauthoryear{Hawkins \bgroup \em et al.\egroup
  }{1980}]{hawkins1980identification}
D~Hawkins, A~Jain, R~Dubes, et~al.
\newblock Identification of outliers chapman and hall.
\newblock {\em London:[Google Scholar]}, 1980.

\bibitem[\protect\citeauthoryear{Henzinger \bgroup \em et al.\egroup
  }{2020}]{outsidethebox19}
Thomas~A. Henzinger, Anna Lukina, and Christian Schilling.
\newblock {\em Outside the Box: Abstraction-Based Monitoring of Neural
  Networks}, volume 325 of {\em Frontiers in Artificial Intelligence and
  Applications}.
\newblock {IOS} Press, 2020.

\bibitem[\protect\citeauthoryear{Ibitoye \bgroup \em et al.\egroup
  }{2019}]{ibitoye2019analyzing}
Olakunle Ibitoye, Omair Shafiq, and Ashraf Matrawy.
\newblock Analyzing adversarial attacks against deep learning for intrusion
  detection in iot networks.
\newblock In {\em 2019 IEEE Global Communications Conference (GLOBECOM)}, pages
  1--6. IEEE, 2019.

\bibitem[\protect\citeauthoryear{Kraft and others}{1988}]{kraft1988software}
Dieter Kraft et~al.
\newblock A software package for sequential quadratic programming.
\newblock DFVLR Obersfaffeuhofen, Germany, 1988.

\bibitem[\protect\citeauthoryear{Mandelbaum and
  Weinshall}{2017}]{mandelbaum2017distance}
Amit Mandelbaum and Daphna Weinshall.
\newblock Distance-based confidence score for neural network classifiers.
\newblock {\em arXiv preprint arXiv:1709.09844}, 2017.

\bibitem[\protect\citeauthoryear{Nassar}{2020}]{nassar-dlh-2020}
Mohamed Nassar.
\newblock {\em Deep Learning Handbook}.
\newblock Zenodo, 2020.
\newblock \url{http://mnassar.github.io/deeplearninghandbook}.

\bibitem[\protect\citeauthoryear{Papadopoulos \bgroup \em et al.\egroup
  }{2021}]{papadopoulos2021launching}
Pavlos Papadopoulos, Oliver Thornewill~von Essen, Nikolaos Pitropakis, Christos
  Chrysoulas, Alexios Mylonas, and William~J Buchanan.
\newblock Launching adversarial attacks against network intrusion detection
  systems for iot.
\newblock {\em Journal of Cybersecurity and Privacy}, 1(2):252--273, 2021.

\bibitem[\protect\citeauthoryear{Pidhorskyi \bgroup \em et al.\egroup
  }{2018}]{pidhorskyi2018generative}
Stanislav Pidhorskyi, Ranya Almohsen, and Gianfranco Doretto.
\newblock Generative probabilistic novelty detection with adversarial
  autoencoders.
\newblock In {\em Advances in neural information processing systems}, pages
  6822--6833, 2018.

\bibitem[\protect\citeauthoryear{Powell}{1994}]{powell1994direct}
Michael~JD Powell.
\newblock A direct search optimization method that models the objective and
  constraint functions by linear interpolation.
\newblock In {\em Advances in optimization and numerical analysis}, pages
  51--67. Springer, 1994.

\bibitem[\protect\citeauthoryear{Price}{2013}]{price2013differential}
Kenneth~V Price.
\newblock Differential evolution.
\newblock In {\em Handbook of optimization}, pages 187--214. Springer, 2013.

\bibitem[\protect\citeauthoryear{Rauber \bgroup \em et al.\egroup
  }{2020}]{rauber2017foolboxnative}
Jonas Rauber, Roland Zimmermann, Matthias Bethge, and Wieland Brendel.
\newblock Foolbox native: Fast adversarial attacks to benchmark the robustness
  of machine learning models in pytorch, tensorflow, and jax.
\newblock {\em Journal of Open Source Software}, 5(53):2607, 2020.

\bibitem[\protect\citeauthoryear{Roberts \bgroup \em et al.\egroup
  }{2019}]{roberts2019bayesian}
Ethan Roberts, Bruce~A Bassett, and Michelle Lochner.
\newblock Bayesian anomaly detection and classification.
\newblock {\em arXiv preprint arXiv:1902.08627}, 2019.

\bibitem[\protect\citeauthoryear{Virtanen \bgroup \em et al.\egroup
  }{2020}]{2020SciPy-NMeth}
Pauli Virtanen, Ralf Gommers, Travis~E. Oliphant, Matt Haberland, Tyler Reddy,
  David Cournapeau, et~al.
\newblock {{SciPy} 1.0: Fundamental Algorithms for Scientific Computing in
  Python}.
\newblock {\em Nature Methods}, 17:261--272, 2020.

\bibitem[\protect\citeauthoryear{Z{\"u}gner \bgroup \em et al.\egroup
  }{2018}]{zugner2018adversarial}
Daniel Z{\"u}gner, Amir Akbarnejad, and Stephan G{\"u}nnemann.
\newblock Adversarial attacks on neural networks for graph data.
\newblock In {\em Proceedings of the 24th ACM SIGKDD International Conference
  on Knowledge Discovery \& Data Mining}, pages 2847--2856, 2018.

\end{thebibliography}
\end{document}